\documentclass{article}

\usepackage{microtype}
\usepackage{graphicx}
\usepackage{subcaption}
\usepackage{xspace}
\usepackage{booktabs} 

\usepackage{hyperref}

\usepackage[preprint]{icml2026}

\usepackage{amsmath}
\usepackage{amssymb}
\usepackage{mathtools}
\usepackage{amsthm}
\usepackage{float}

\usepackage[capitalize,noabbrev]{cleveref}

\theoremstyle{plain}

\theoremstyle{definition}

\theoremstyle{remark}

\usepackage[textsize=tiny]{todonotes}

\icmltitlerunning{Mitigating Subject Dependency in EEG Decoding with Subject-Specific Low-Rank Adapters}

\usepackage{comment}
\usepackage{multirow}
\usepackage{graphicx} 
\usepackage{placeins}
\newcommand{\method}{SuLoRA\xspace} 
\usepackage{enumitem}

\begin{document}

\twocolumn[
  \icmltitle{Mitigating Subject Dependency in EEG Decoding \\
with Subject-Specific Low-Rank Adapters}



  \icmlsetsymbol{equal}{*}

  \begin{icmlauthorlist}
    \icmlauthor{Timon Klein}{OVGU}
    \icmlauthor{Piotr Minakowski}{OVGU}
    \icmlauthor{Sebastian Sager}{OVGU,MPI}
    \icmlauthor{Steffen Schotthöfer}{ORNL}
  \end{icmlauthorlist}

  \icmlaffiliation{OVGU}{Department of Mathematics; Otto von Guericke University Magdeburg; 39106 Magdeburg; Germany.}
  \icmlaffiliation{MPI}{Max Planck Institute for Dynamics of Complex Technical Systems; 39106 Magdeburg; Germany.}
  \icmlaffiliation{ORNL}{Computer Science and Mathematics Division; Oak Ridge National Laboratory; Oak Ridge, TN 37831; USA}
  \icmlcorrespondingauthor{Timon Klein}{timon.klein@ovgu.de}
  \icmlkeywords{Machine Learning, ICML}
  \vskip 0.3in
]



\printAffiliationsAndNotice{}  

\begin{abstract}
Subject-specific distribution shifts represent a fundamental obstacle to developing foundation models for brain decoding. 
We propose the Subject-Specific Low-Rank Adapter (SuLoRA), a drop-in replacement for standard linear or convolutional layers that captures inter-subject variability by decomposing weights into a shared, subject-invariant component and a lightweight, low-rank correction unique to each subject. 
This explicit separation enables existing architectures to become robust to subject shifts without architectural redesign. 
We evaluate SuLoRA on MEG speech perception and EEG motor imagery tasks across CNN and transformer architectures. 
In the speech decoding task, SuLoRA exceeds the baseline performance with half of the parameters. 
On motor imagery dataset, SuLoRA outperforms both subject-agnostic models and independently trained subject-specific models. SuLoRA offers a practical path towards effective cross-subject foundation models for brain signal applications.

\end{abstract}

\section{Introduction}
Developing a foundation model for decoding complex brain signals such as Electroencephalography (EEG) and magnetoencephalography (MEG) is a central challenge in neuroscience and machine learning \cite{aristimunha2025neuripsEEGchallenge}. 
Progress is slowed by the inherent properties of EEG data: low signal-to-noise ratio, non-stationarity, and, most critically, high inter-subject variability \cite{wei20222021beetlcompetitionadvancing}. 
Neural patterns vary so widely between individuals and even within the same individual over time. The data are effectively fragmented, amplifying the problem of data scarcity \cite{banville2025scalinglawsdecodingimages, bomatter2025limitedparticipantdiversityimpeding, Schmid2024}. 
Consequently, large models trained on pooled data from many 
subjects often fail to outperform simpler models tailored to 
individuals, slowing progress towards generalizable solutions.

This situation contrasts sharply with fields of natural language processing or computer vision. 
Although these fields have also relied on domain-specific 
architectures, the Transformer \cite{vaswani2023attentionneed} 
provided a universal paradigm that unlocked predictable scaling 
laws, allowing massive foundation models to emerge \cite{hoffmann2022Chinchilla, zhai2022scalingvisiontransformers}. 
Brain decoding has yet to benefit from similar scaling. The 
challenge of cross-subject generalization has kept effective 
dataset sizes small, preventing the successful application of 
large-scale, general-purpose architectures. 
As a result, the field remains dominated by specialized models that have not yet converged to a standard, foundational approach \cite{aristimunha2025neuripsEEGchallenge, Guetschel2024_Review_BCI}.

To bridge this gap, we address the problem of subject-specific 
distribution shifts, which we identify as the primary obstacle 
to building foundation models for multi-subject brain-signal applications.
We propose the Subject-Specific Low-Rank Adapter (\method), an 
approach that decomposes the parameters of a neural network into 
two subsets: a set of base weights shared across all subjects 
and a subject-specific, low-rank correction that efficiently 
captures individual variability. 
The {\method} module is designed as a drop-in replacement for standard linear or convolutional layers, making it universally applicable to a wide range of model architectures with applications in EEG/MEG decoding and beyond.

Our contributions read as follows:
\begin{itemize}[leftmargin=*, noitemsep, topsep=0pt]
\item We introduce {\method}, a novel adaptive layer that explicitly separates shared knowledge from subject-specific signatures, designed as a drop-in replacement for linear and convolutional layers in any architecture.
\item We provide qualitative evidence through embedding visualization that {\method} successfully disentangles subject-invariant neural patterns from individual signatures.
\item We demonstrate through comprehensive evaluation on speech perception (MEG) and motor imagery (EEG) tasks that models equipped with {\method} outperform: (i) subject-agnostic models, (ii) independently trained subject-specific models, and (iii) independent subject-specific LoRA models.
\end{itemize}

Code\footnote{\hyperlink{https://github.com/timonkl/SubjectConditionedLayer}{https://github.com/timonkl/SubjectConditionedLayer}} and data are publicly available.

\section{Related Work}
Scaling laws in natural language processing and computer vision have established a reliable paradigm for improving predictive performance by jointly increasing model size, training compute, and dataset scale \cite{kaplan2020scalinglawsneurallanguage, hoffmann2022Chinchilla, zhai2022scalingvisiontransformers}. 
In contrast, recent work on scaling laws for neural decoding shows that performance improves with increased training data per subject, but does not scale proportionally with the number of subjects \cite{banville2025scalinglawsdecodingimages}. 
This presents a fundamental obstacle to developing large-scale, multi-subject foundation models for brain decoding, where aggregating data across individuals is both necessary and desirable.
Despite these challenges, Leave-One-Subject-Out (LOSO) benchmarks \cite{Wang_2020Physionet, Kwon_LOSO, zhang2025} demonstrate that cross-subject transfer is generally possible, although with notable performance degradation. 
Taken together, these findings suggest that brain decoding representations comprise both transferable subject-agnostic components and subject-specific components that limit performance gains when scaling across subjects.

To address inter-subject variability in brain decoding, a wide range of approaches has been proposed.
Early methods introduce subject-dependent bias terms or linear transformations applied to learned features \cite{Ma2019}. 
Recent studies investigate inter-subject variability via explicit adaptation mechanisms, 
such as subject-conditioned batch normalization and adaptive feature modulation. In particular, 
\cite{Liang2024} learn a shared subject-invariant feature space while applying lightweight 
subject-specific transformations to correct individual distortions, enabling cross-subject decoding using a small amount of target-subject data.

More recent work in speech decoding \cite{lévy2025braintotextdecodingnoninvasiveapproach, dascoli2024decodingindividualwordsnoninvasive, shapovalenko2025} introduces a subject-specific linear layer applied early in the network.
This layer performs a channel-wise transformation that aligns subject-specific representations into a shared latent space while keeping the remaining parameters shared across subjects.
However, this approach introduces a relatively large number of additional parameters that grow linearly with the number of subjects.
Ablating these subject-specific layers leads to significant accuracy drops \cite{defossez2023decoding}, underscoring the importance of explicitly modeling subject-dependent effects.

Our \method\ directly addresses inter-subject variability by explicitly learning both shared and subject-specific representations.
It introduces a set of global parameters augmented with lightweight, subject-specific corrections.
This design is inspired by nonlinear mixed-effects models \cite{lindstrom1990nonlinear, pinheiro2000mixed}, which jointly model population-level effects and parsimonious individual-level deviations via random effects.
The approach also naturally aligns with multi-task learning (MTL) \cite{caruana1997multitask, Evgeniou2004_multi_task_learning, Ruder17a_Multi_Task_DNN}, where shared structure across related tasks is leveraged while allowing task-specific variation.

Conceptually, our approach is related to Mixture-of-Experts (MoE) models \cite{jiang2024mixtralexperts, dai2024deepseekmoeultimateexpertspecialization, riquelme2021scalingvisionsparsemixture}, which decompose computation across multiple expert networks and use a gating mechanism to select experts for each input \cite{Jacobs1991}. 
Sparsely gated MoE variants enable scaling model capacity without proportional increases in computational cost by activating only a subset of experts per example \cite{ShazeerSparselyGatedMoE, fedus2022switchtransformersscalingtrillion}.
MoE architectures have been adapted for brain decoding: EEGMamba \cite{gui2024eegmambabidirectionalstatespace} employs a task-aware learned gating network together with a universal expert to capture shared EEG structure, while MoGE \cite{liu2024moge} and VMoGE \cite{ding2025variationalmixturegraphneural} explore graph-based experts that assign channels or frequency bands to specialized experts.
In brain decoding, subject identity provides a deterministic and known source of variation, acting as a fixed gating signal that routes each input to its corresponding expert pathway.
Unlike conventional MoE models that rely on learned routing policies, our approach exploits this prior structure directly, simplifying the architecture while explicitly modeling the source of inter-subject variability that limits cross-subject scaling.

\begin{figure*}[t!]
  \centering
  \includegraphics[width=0.8\linewidth]{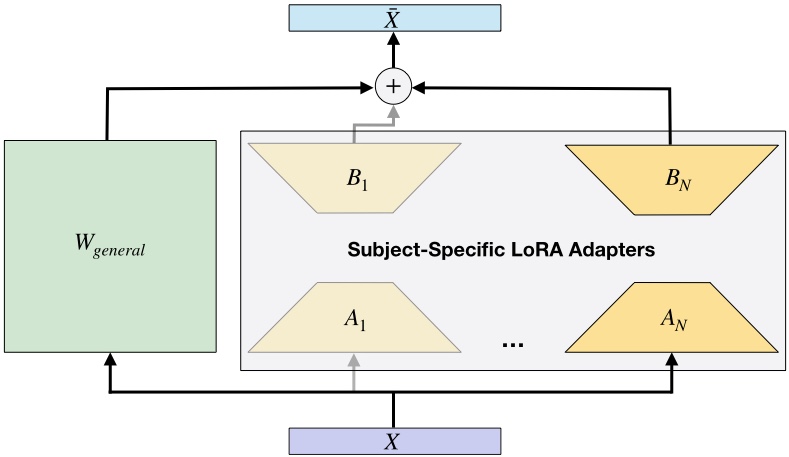}
  \caption{\textbf{{\method} architecture}. The input $X$ is processed by a shared weight matrix $W_{\text{general}} \in \mathbb{R}^{m \times n}$ capturing subject-invariant patterns. Each subject s has a dedicated low-rank adapter, parameterized by matrices $A_s \in \mathbb{R}^{n \times r}$ and $B_s \in \mathbb{R}^{r \times m}$, which computes a subject-specific correction added to the shared output. Unlike standard MoE models, {\method} uses known subject identity to deterministically select the appropriate adapter at both training and inference.}
  \label{fig:model}
\end{figure*}

Our implementation leverages a shared full-rank expert that is always active, together with subject-specific LoRA modules.
LoRA \cite{hu2021lora}, originally introduced as a parameter-efficient fine-tuning method that adds low-rank updates to pretrained weights, has been successfully applied across modalities and tasks, offering strong empirical performance with minimal memory and compute overhead.
Prior work has explored design choices such as rank selection \cite{ZangrandoGeometry_aware, schotthöfer2024federateddynamicallowranktraining} and robust compression strategies \cite{schotthöfer2025dynamicallowrankcompressionneural}.
In our framework, LoRA serves as the mechanism for implementing subject-specific corrections, enabling fine-grained adaptation to individual subjects while tightly controlling the number of additional parameters.
It has also been shown that LoRA enables parameter-efficient 
fine-tuning of large brainwave models, achieving competitive EEG 
decoding performance with minimal trainable parameters \cite{lee2025loraeeg}.

\section{On the Difficulty of Multi-Subject Training} \label{sec:theory}
In this section, we focus solely on a classification problem in 
which data is collected from multiple subjects. Since the signal 
contains subject-specific characteristics, our goal is to learn 
a general model that captures class-discriminative structures 
across subjects.
Let $\mathcal{X}_s = \{x_{sj}\}_{j=1}^{M_s}$ and $\mathcal{Y}_s = \{y_{sj}\}_{j=1}^{M_s}$ contain subject-specific features and labels, respectively, for subjects $s \in \{1, \dots , N\}$. In short, we write $\mathcal{D} = (\mathcal{X}, \mathcal{Y}) = \bigcup_{s=1}^N \mathcal{D}_s = \bigcup_{s=1}^N (\mathcal{X}_s, \mathcal{Y}_s)$.

A single linear layer trained for all subjects implicitly estimates $p(\mathcal{Y} | \mathcal{X} )$, but cannot properly capture subject-specific features because it lacks a mechanism to adapt its weights per subject. It can only represent the shared average structure.
We address this with a probabilistic framework that models subject-specific effects. Our main goal is to learn the subject-independent classifier $p(\mathcal{Y} | \mathcal{X} )$ that for a given $\mathcal{D}$ marginalizes over $s$,
\begin{align}  
p(\mathcal{Y}\mid \mathcal{X}) = \textstyle\sum_{s=1}^N  p(\mathcal{Y}\mid \mathcal{X}, s). 
\end{align}

Further, assume that we have latent random variables $\mathcal{Z} = \xi(\mathcal{X},s)$ as intermediates on the map from $\mathcal{X}$ to $\mathcal{Y}$, where we define $\xi$ as a transformation to latent representation. Then we can write
\begin{align}\label{eq:prob2}
p(\mathcal{Y} \mid \mathcal{X}, s) = p(\mathcal{Y} \mid \mathcal{Z} ,s) \; p( \mathcal{Z} \mid \mathcal{X}, s).
\end{align}

Concerning the first term of Equation~\eqref{eq:prob2} we assume that
$$p(\mathcal{Y} \mid \mathcal{Z} ,s) = p(\mathcal{Y} \mid \mathcal{Z}),$$
which states that once the latent representation $\mathcal{Z}$ is known, the participant index $s$
 and the datapoint $x_{sj}$ provide no additional information about the label
$y_{sj}$.  This conditional independence assumption implies that
$\mathcal{Z}$ serves as a sufficient statistic of $x_{sj}$ to predict $y_{sj}$.

The second term of Equation~\eqref{eq:prob2}, $p(\mathcal{Z} \mid \mathcal{X}, s)$, describes how the latent representation is learned. Specifically,  
\begin{align}
    \xi(\mathcal{X},s) \sim p(\mathcal{Z} \mid \mathcal{X}, s),
\end{align}
i.e., $\xi$ is treated as a stochastic latent variable in the representation space $\mathcal{Z}$, sampled conditionally on the input $\mathcal{X}$ and the participant index $s$.
The subject-specific map $\xi(\mathcal{X},s)$ is parameterized by a set of general weights and a subject-specific low-rank correction.

\section{Method}

\subsection{Subject-Specific
Low-Rank Adapter}\label{sec:sub_lin_layer}

We propose a novel subject-aware layer for multi-subject machine learning tasks as
\begin{align}
    \bar{X} = 
    \sigma\left(X W_{\text{general}}^\top + \textstyle\sum_{s=1}^{N} (M_s \cdot X) W_{s}^\top \right), \label{eq:subject-layer}
\end{align}
where $W_{\text{general}} \in \mathbb{R}^{m \times n}$ and $W_s \in \mathbb{R}^{m \times n}$ are shared and subject-specific weight matrices, respectively.
The mask matrices $M_s$ with binary entries extract subject-$s$-specific samples from the batch data $X$. 
Note that this approach may lead to a reduction in batch sizes for subject-dependent weights $W_{s}$, if the number of subjects $N$  increases. 
As a remedy, adapting the total batch size was proposed in \cite{ShazeerSparselyGatedMoE}.

\subsection{Low Rank approximation for Subject-Dependent Weights}\label{sec:lowRank}

Low-rank adaptation (LoRA) \cite{hu2021lora} is a well-established method for fine-tuning models in few-shot settings. 
Compared to full-rank training, LoRA shows superior performance on limited data \cite{schotthoefermomentum-based}, while also substantially reducing the number of training parameters and the associated memory consumption. 
We define subject-specific weights as low-rank matrix decompositions 
\begin{align}\label{eq:lowRankSub}
    W_{s}^\top := A_s B_s, 
\end{align}
with $A_s \in \mathbb{R}^{n \times r}, \; B_s \in \mathbb{R}^{r \times m}$ and rank $r$. 
Unlike standard LoRA, we initialize $A_s \sim \mathcal{N}\left(0, \frac{2}{r}\right)$ and $B_s \sim \mathcal{N}(0, 0.01^2)$ for all subjects $s$. Initializing $B_s = 0$, as in standard LoRA, led to reduced performance, likely due to vanishing gradients through $A_s$ in early training.

\subsection{Scaling of Subject-Specific Adapters}
Similar to the original LoRA, we introduce a scaling of $W_{s}$ with 
$\frac{\alpha}{r}$, where $\alpha$ is a hyperparameter. 
This scaling adjusts the effective learning rate of the adapter matrices relative to the general weights. 
For $\alpha < r$, the adapters are updated conservatively, encouraging reliance on $W_{\text{general}}$, 
in contrast to $\alpha > r$, where the adapters are emphasized.

\subsection{Regularization of Subject-Specific Adapters}
A central challenge is to ensure that the general weight matrix $W_{\text{general}}$ learns shared patterns while the subject-specific weights $W_{s}$ only capture individual deviations. 
To prevent $W_{s}$ from learning general representations, we employ a regularization strategy. We constrain the complexity of the subject-specific corrections by enforcing a relatively small rank $r$, as defined in Equation~\ref{eq:lowRankSub}. 
See Tables~\ref{tab:hyperparametersEEGNeX} and \ref{tab:hyperparametersPBT} for exact values. 

The weights $W_{s}$ could also be regularized by adding a weight regularization term (e.g., an $L_2$ penalty) to the loss function. 
We did not pursue this here, opting instead to demonstrate the capabilities of the LoRA only approach.

\section{Decoding Tasks and Datasets} 
\label{sec:data}
We evaluate our approach on two distinct neural decoding tasks: speech perception from MEG recordings and motor imagery classification from EEG data.

\subsection{Speech Perception}\label{sec:speachtask}
We employ the MEG dataset \cite{Gwilliams2023}, a high-quality MEG dataset for natural speech processing. 
The data set consists of 208-channel MEG recordings from 27 English-speaking participants who listened to four fictional stories from the Masc corpus \cite{ide-etal-2010-manually}. 
The total recording duration is 56.2h. For each subject, data were collected in two 1-hour sessions, although the second session is missing for five subjects.
This yields 4417 training segments and 1363 test segments with a vocabulary of 1810 and 846 unique words, respectively, with 64\% word overlap between splits.
Following \cite{defossez2023decoding}, we use the same data split and preprocessing procedure, and formulate the speech-perception decoding task as a segment retrieval problem: given a short window of MEG signal, the model must identify the corresponding speech segment from a large set of candidate audio segments.

\subsection{Motor Imagery}\label{sec:motorimaginarytask}
To complement retrieval-based speech decoding with Brain Computer Interface (BCI) classification, we additionally consider two datasets from BCI Competition IV \cite{BCIC_IV}.
The BCI Competition IV 2a dataset \cite{BCIC_IV2a} consists of 22-channel EEG recordings from nine subjects. For each subject, data were collected in two sessions (training and evaluation), with 288 trials per session. The labels are balanced across four motor imagery classes: left hand, right hand, foot, and tongue.
The BCI Competition IV 2b dataset \cite{BCIC_IV2b, BCIC_IV2b2} consists of 3-channel EEG recordings from nine subjects. For each subject, data were collected across five sessions, three for training and two for evaluation, totaling between 680 and 760 trials. The labels are balanced between the classes: left-hand and right-hand.

\section{Architectures and Training Objectives}
\label{sec:benchmark_architectures}

To demonstrate the generality of \method as a drop-in architectural component, we benchmark it across three distinct modeling setups: a convolutional network, a vision transformer-based architecture, and a contrastively trained representation-learning pipeline.

\subsection{BrainMagick}
\label{sec:brainmagick}
BrainMagick \cite{defossez2023decoding} addresses the challenging task of decoding speech from non-invasive MEG and EEG recordings. 
The architecture consists of two components: a Brain module and a Speech module.
The Brain module comprises a spatial attention mechanism that encodes MEG/EEG sensors, a subject-specific linear layer, and a stack of convolutional blocks. 
The Speech module is a frozen wav2vec~2.0 speech encoder \citep{baevski2020wav2vec}.
The Brain module is trained using a CLIP (Contrastive Language–Image Pre-Training) loss \cite{radford2021clip} to align latent brain representations with speech representations.
A critical design choice in BrainMagick is the use of a subject-specific layer to handle substantial inter-individual variability in MEG/EEG signals. 
This layer accounts for approximately 2M of the 9.5M (for 27 subjects) trainable model parameters and scales linearly with the number of subjects.
We replaced the subject-specific layer of the BrainMagick with \method.

\subsection{EEGNeX}
\label{sec:eegnex}
EEGNeX~\citep{Chen2024EEGNeX} represents the state-of-the-art among compact, convolution-only architectures for motor imagery classification, that refines the widely-adopted EEGNet~\cite{Lawhern2018EEGNet}. The network consists of five sequential blocks. Blocks~1 and~2 apply temporal convolutions with batch normalization to learn frequency-selective filter banks, operating analogously to learned Finite Impulse Response filters. Block~3 performs depthwise spatial convolution across electrode channels with max-norm constraints, followed by ELU activation, average pooling, and dropout. Blocks~4 and~5 introduce dilated temporal convolutions to expand the receptive field while maintaining parameter efficiency, with the final block including additional pooling and flattening operations. A linear layer constrained by the maximum norm serves as the classifier.
We integrate \method by replacing standard convolutional layers  while preserving depthwise spatial convolution in Block~3.

\subsection{Patched Brain Transformer}
\label{sec:pbt}
The Patched Brain Transformer  (PBT)~\citep{Klein2025PBT} adapts the Vision Transformer \cite{dosovitskiy2021imageworth16x16words} architecture for EEG decoding, that reflect recent trends in the field \cite{Kostas2021BENDR, jiang2024LaBraM}. Raw EEG signals are segmented into fixed-duration patches, which are linearly projected into token embeddings. 
A learned positional embedding encodes both spatial (electrode) and temporal (time window) information, enabling the model to flexibly handle varying sensor configurations and recording durations. 
The core architecture follows the standard Transformer encoder design.
We apply \method to the linear projections within the Transformer encoder, decomposing the query, key, value, and feed-forward weight matrices into shared and subject-specific components.

\section{Results}
We evaluate our method on diverse neural signal decoding tasks: speech perception from MEG 
recordings, and motor imagery classification from EEG, demonstrating the generality of our approach 
across different brain recording modalities and cognitive paradigms.

\begin{table}[h]
  \caption{\textbf{Speech decoding performance} on the \cite{Gwilliams2023} MEG dataset. We replace the subject-specific layer from BrainMagick~\citep{defossez2023decoding} (Baseline) with \method at varying ranks $r$ and scaling factors $\alpha$. ``BL w/o Sub.-Layer" refers to the Baseline where the subject-specific layer is replaced with the same linear layer for all subjects. The parameter count indicates the number of parameters in the subject-specific layer for each respective method. \method with $r=8$ nearly matches the Baseline performance while reducing parameters by an order of magnitude. \method with $r=32$ improves upon Baseline results using only half the number of parameters.}
  \label{tab:BrainMagick}
  \centering
{ 
\renewcommand{\arraystretch}{1.2}
  \resizebox{\columnwidth}{!}{
  \begin{tabular}{lccccc}
    \toprule
 & \textbf{Method} & \makebox[0pt][c]{\textbf{subject-specific}} & \multicolumn{3}{c}{\textbf{Accuracy [\%] $\pm$ Std. [\%]}} \\
\cmidrule(lr){4-6}
& & \textbf{\#Params} & \textbf{Top 1} & \textbf{Top 5} & \textbf{Top 10} \\
        \midrule
        \multirow{17}{*}{\rotatebox{90}{\textbf{\method}}\hspace{-0.7em}}
    & r=4, $\alpha$=4          &   \phantom{1,}131k   &    37.56 $\pm$ 0.67     &   58.72 $\pm$ 0.52   &   67.06 $\pm$ 0.54 \\
    & r=4, $\alpha$=8          &   \phantom{1,}131k  &     38.85 $\pm$ 0.33   &   60.17 $\pm$ 0.45 &  68.52 $\pm$ 0.39 \\
    & r=4, $\alpha$=12 &   \phantom{1,}131k   & 39.04 $\pm$  0.20 & 60.36 $\pm$ 0.21 & 68.77 $\pm$  0.36 \\
    & r=8, $\alpha$=8          &   \phantom{1,}190k   &   39.82 $\pm$ 0.72   &   61.03 $\pm$ 0.68   &   69.39 $\pm$ 0.62 \\
    & r=8, $\alpha$=16         &   \phantom{1,}190k   &   39.61 $\pm$ 0.48   &   61.02 $\pm$ 0.27   &   69.40 $\pm$ 0.37 \\
    & r=8, $\alpha$=24    &   \phantom{1,}190k   &  40.32  $\pm$  0.50  & 61.5 $\pm$ 0.23   &   69.79 $\pm$ 0.29  \\
    & r=16, $\alpha$=16        &   \phantom{1,}306k   &    40.50 $\pm$ 0.27   &   61.77 $\pm$ 0.19  &   69.97 $\pm$ 0.22 \\
    & r=16, $\alpha$=32        &   \phantom{1,}306k  &     40.50 $\pm$ 0.27  &   61.86 $\pm$ 0.32   &   70.09 $\pm$ 0.29\\
    & r=16, $\alpha$=48        &   \phantom{1,}306k  &   40.68 $\pm$ 0.51   &    61.94 $\pm$ 0.39   & 70.20  $\pm$ 0.36 \\
    & r=32, $\alpha$=32        &   \phantom{1,}540k   &  40.95$\pm$ 0.54  &   62.22 $\pm$ 0.64   &   70.34 $\pm$ 0.47 \\
    & r=32, $\alpha$=64        &   \phantom{1,}540k   &   41.65 $\pm$ 0.76   &   62.63 $\pm$ 0.84   &   70.76 $\pm$ 0.76 \\
    & r=32, $\alpha$=96        &   \phantom{1,}540k   &  41.44 $\pm$ 0.32  &    62.67 $\pm$  0.31  &   70.60 $\pm$ 0.18 \\
    & r=64, $\alpha$=64        &   1,006k             &   40.97 $\pm$ 0.44   &   62.15 $\pm$ 0.49   &   70.30 $\pm$ 0.49 \\
    & r=64, $\alpha$=128        &   1,006k             &   41.74 $\pm$ 0.49   &   62.90 $\pm$ 0.43   &   70.98 $\pm$ 0.45 \\
    & r=64, $\alpha$=192        &   1,006k             &  \textbf{41.94 $\pm$ 0.92}      &   \textbf{63.20 $\pm$ 0.65}     &   \textbf{71.18 $\pm$ 0.61} \\
    & r=64, $\alpha$=256        &   1,006k             &   41.81 $\pm$ 0.21   &   63.00 $\pm$ 0.11   &   70.95 $\pm$ 0.21 \\
    & r=96, $\alpha$=192        &   1,473k                 &  41.62 $\pm$ 0.13     & 62.73 $\pm$ 0.17  & 70.85 $\pm$ 0.16 \\
    \midrule
    & Baseline                 &   1,968k             &  40.60 $\pm$ 0.18  &   61.84 $\pm$ 0.21  &   70.00 $\pm$ 0.21 \\
    \makebox[0pt][l]{BL w/o Sub.-Layer}    &       &   \phantom{1,1} 73k             & 19.23 $\pm$  0.41 &    36.69 $\pm$ 0.54  &   45.48 $\pm$ 0.69 \\
    
    \bottomrule
  \end{tabular}
  }
}
  
\end{table}

\begin{table*}[ht]
    \caption{\textbf{Zero-Shot.} Prediction accuracy (\%) under Leave-One-Subject-Out evaluation of EEGNeX \cite{Chen2024EEGNeX} on BCI Competition IV-2a using \method, compared to the subject-agnostic baseline. \method\ achieves competitive accuracy relative to the baseline when relying solely on the general weights, indicating that general EEG features are effectively captured by these weights (highest values in bold). \textbf{Transferred Adapters.} Prediction accuracy (\%) under Leave-One-Subject-Out evaluation when transferring an adapter trained on one training subject to predict the held-out subject. Adapter $i$ corresponds to subject $i$; therefore, diagonal entries are omitted. Transferred adapters outperform the baseline, enabling a form of few-shot learning by either selecting an appropriate adapter or using it as an initialization for LoRA fine-tuning. (Values shown in bold when higher than the Zero-Shot results for \method and baseline.)}
  \label{tab:LOSO}
  \centering
  \resizebox{\textwidth}{!}{
  \renewcommand{\arraystretch}{1.1}
  \begin{tabular}{cccccccccc}
    \toprule
     \textbf{Method} & \multicolumn{9}{c}{\textbf{Leave Out Subject}} \\
     &  \multicolumn{9}{c}{Accuracy [\%] $\pm$ Std. [\%]}\\
                \midrule
         &       \multicolumn{9}{c}{\textbf{Zero-Shot}} \\
                    & \textbf{S1} & \textbf{S2} & \textbf{S3} & \textbf{S4}& \textbf{S5} & \textbf{S6} & \textbf{S7} & \textbf{S8} & \textbf{S9} \\
EEGNeX & \textbf{62.27 $\pm$ 2.56} & \textbf{26.97 $\pm$ 0.86} & 52.31 $\pm$ 3.46& \textbf{36.92 $\pm$ 2.80} & 24.77 $\pm$ 0.33 & \textbf{28.36 $\pm$ 1.40} & 28.70$\pm$ 1.28 & 54.52 $\pm$ 2.25 & 44.21 $\pm$ 2.64 \\ 
   EEGNeX \method & 37.04 $\pm$ 4.02 & 25.23 $\pm$ 0.16 & \textbf{56.37 $\pm$ 4.52} & 32.76 $\pm$ 0.16 & \textbf{25.12 $\pm$ 0.16} & 28.13 $\pm$ 1.50 & \textbf{28.93 $\pm$ 0.59} & \textbf{58.22 $\pm$ 1.99} & \textbf{45.14 $\pm$ 2.04}\\   
        \midrule
\textbf{Adapter} & \multicolumn{9}{c}{\textbf{Transferred Adapters}} \\
           & \textbf{S1} & \textbf{S2} & \textbf{S3} & \textbf{S4}& \textbf{S5} & \textbf{S6} & \textbf{S7} & \textbf{S8} & \textbf{S9} \\
1 & n/a              & 28.01 $\pm$ 0.82 & \textbf{67.94 $\pm$ 3.23} & 34.14 $\pm$ 1.61 & 24.65 $\pm$ 0.28 & 29.28 $\pm$ 1.89 & 34.49 $\pm$ 1.93 & 46.41 $\pm$ 4.03 & 34.95 $\pm$ 0.87 \\
2 & 45.83 $\pm$ 3.71 & n/a & 28.59 $\pm$ 1.66 & 31.71 $\pm$ 2.05 & \textbf{28.70 $\pm$ 2.90} & 26.39 $\pm$ 0.57 & 16.43 $\pm$ 2.36 & 31.25 $\pm$ 3.97 & 24.19 $\pm$ 1.66 \\
3 & 61.81 $\pm$ 3.00 & 28.12 $\pm$ 0.85 & n/a & 27.43 $\pm$ 0.98 & 26.97 $\pm$ 1.56 & 31.14 $\pm$ 1.45 & 29.98 $\pm$ 1.56 & 48.84 $\pm$ 5.42 & 39.01 $\pm$ 3.70 \\
4 & 47.34 $\pm$ 5.57 & 22.92 $\pm$ 1.02 & 33.91 $\pm$ 5.70 & n/a & 26.27 $\pm$ 1.31 & \textbf{31.60  $\pm$ 2.32} & 22.80 $\pm$ 2.54 & 54.86 $\pm$ 2.60 & 34.26 $\pm$ 2.01 \\
5 & 50.00 $\pm$ 6.13 & 27.55 $\pm$ 2.87 & 39.70 $\pm$ 6.11 & 32.87 $\pm$ 1.99 &  n/a& 25.46 $\pm$ 0.33 & \textbf{31.60 $\pm$ 1.42} & 45.72 $\pm$ 1.61 & 33.45$\pm$ 3.03 \\
6 & 36.46 $\pm$ 0.29 & 25.46 $\pm$ 2.27 & 35.88 $\pm$ 6.29 & 37.96 $\pm$ 0.16 & 25.00 $\pm$ 0.29 & n/a & 30.21 $\pm$ 1.98 & 45.95 $\pm$ 2.87 & 43.52 $\pm$ 4.61 \\
7 & 52.43 $\pm$ 3.00 & 24.65 $\pm$ 2.14 & 45.95 $\pm$ 6.12 & 26.62 $\pm$ 0.87 & 25.35 $\pm$ 0.00 & 26.62 $\pm$ 1.61 & n/a & 32.18 $\pm$ 0.91 & 32.18 $\pm$ 3.99 \\
8 & 54.51 $\pm$ 2.88 & \textbf{29.05 $\pm$ 2.87} & 59.84 $\pm$ 5.67 & \textbf{43.06 $\pm$ 2.42} & 24.31 $\pm$ 0.57 & 28.94 $\pm$ 2.01 & 28.12 $\pm$ 3.07 & n/a & 45.14 $\pm$ 2.47 \\
9 & 29.05 $\pm$ 0.59 & 25.35 $\pm$ 0.28 & 31.25 $\pm$ 1.99 & 26.27 $\pm$ 1.56 & 25.00 $\pm$ 0.00 & 28.36 $\pm$ 1.61 & 23.26 $\pm$ 2.47 & 41.09 $\pm$ 4.13 & n/a \\
    \bottomrule
  \end{tabular}
  }
\end{table*}

\subsection{Application to Speech Decoding}
To demonstrate that \method\ can serve as an efficient drop-in replacement, we reproduced the speech decoding results from ~\citep{defossez2023decoding} on the MEG dataset~\cite{Gwilliams2023} (see Sec. \ref{sec:speachtask}).
We used the publicly available BrainMagick GitHub repository~\citep{brainmagick_code} with identical hyperparameters and random seeds \{2036, 2037, 2038\}.
Unlike the original work, which used gradient accumulation across two (Nvidia~V100) GPUs, we trained the model on a single (Nvidia~H100) GPU.
Since contrastive losses draw negative samples only within-GPU batches, 
our single-GPU configuration provides access to the full batch of negatives.

In Table~\ref{tab:BrainMagick}, we compare \method with the BrainMagick model (Baseline), that achieves $40.60\%$ a top-1 accuracy (top-10: $70.00\%$) using 1,968k parameters in the subject-specific layer.
Replacing the subject-specific layer with the same linear layer for all subjects, causes performance to collapse to top-1 accuracy of $19.23\%$ (top-10: $45.48\%$), underscoring the importance of modeling inter-subject variability in neural decoding tasks.

\method provides a favorable accuracy-efficiency trade-off across the parameter spectrum. 
For $r=8$ and $\alpha=24$, it closely matched the baseline accuracy of $40.32\%$ top-1 (top-10: $69.79\%$) with only 190k parameters, a reduction of an order of magnitude compared to the baseline.
Peak performance is achieved at $r=64$ with $\alpha=192$, reaching $41.94\%$ top-1, $63.20\%$ top-5, and $71.18\%$ top-10 accuracy. Notably, this configuration still uses roughly half the parameters of the baseline (1,006k vs.\ 1,968k). Further increasing rank does not improve performance.
The scaling factor $\alpha$ plays a crucial role in performance. For a fixed rank, we observe consistent improvements when increasing $\alpha$ up to $\alpha=3r$, after which performance plateaus or slightly decreases.

\subsection{Leave-One-Subject-Out}

\begin{figure*}[ht]
    \centering
    \includegraphics[width=\textwidth]{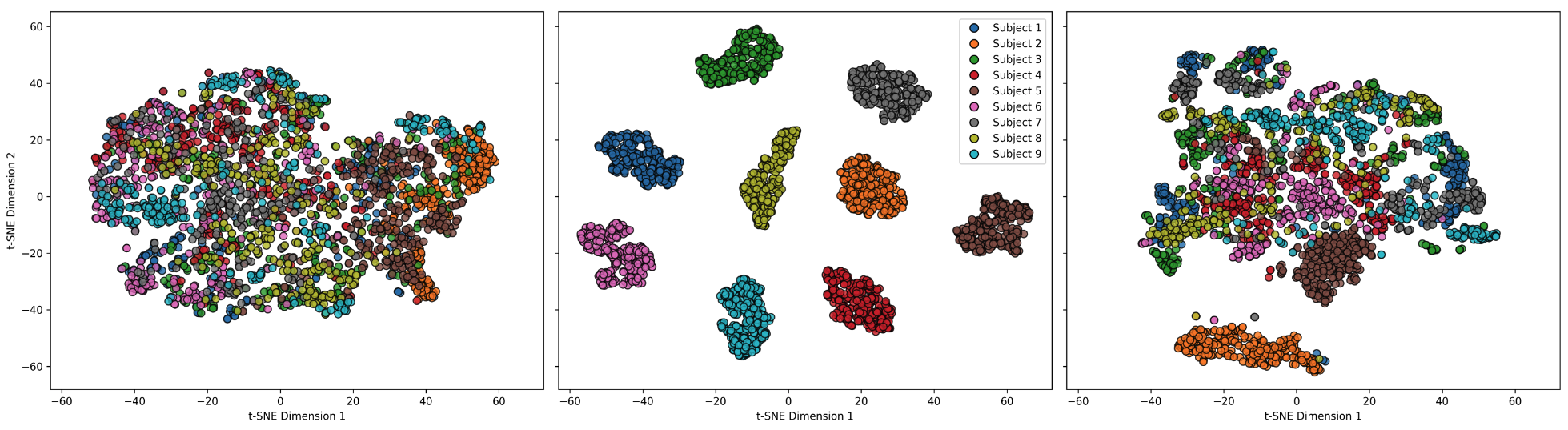}
    \vspace{0.5cm}
    \includegraphics[width=\textwidth]{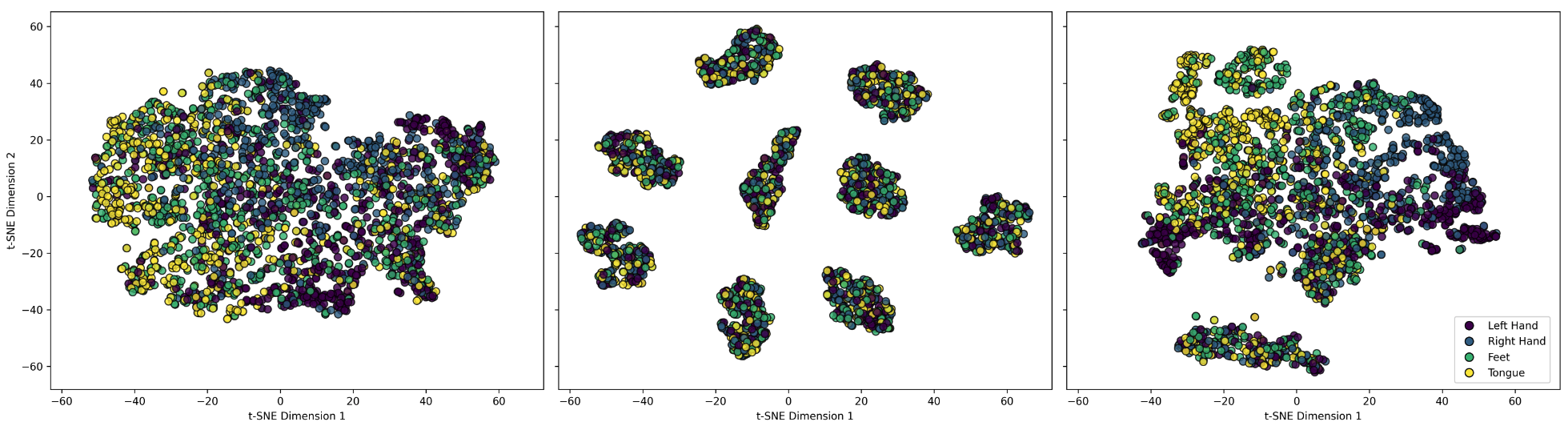}
    \caption{
    \textbf{Latent Representations Colored by Subject ID vs. Target Labels.}
    The top row compares t-SNE projections of representations derived from \textbf{(Left)} the shared weights ($W_{\text{general}}$), \textbf{(Middle)} the subject-specific weights ($W_{s}$), and \textbf{(Right)} the fused representations from the complete {\method}. 
    In the top row, all points are colored by subject ID.
    The bottom row presents the same data points, but colored according to their ground-truth target labels. 
    This comparison clearly shows that the sub-clusters formed by our full model (top right) correspond directly to the target classes (bottom right), demonstrating that the model learns discriminative, task-relevant features within each subject’s representation space.
    All representations are obtained from the classification token predicted by the Patched Brain Transformer, with a linear classification head used for visualization clarity.
    }
    \label{fig:tSNE}
\end{figure*}

Table~\ref{tab:LOSO} reports prediction accuracy under a Leave-One-Subject-Out (LOSO) evaluation on the BCI Competition IV-2a dataset using EEGNeX.
In each fold, one subject is held out for testing while the model is trained on the remaining eight subjects.
We compare a subject-agnostic baseline, \method in a zero-shot setting, and transferred adapters trained on individual source subjects.

In the zero-shot regime, predictions using \method rely solely on the shared, subject-independent weights.
\method achieves accuracy comparable to the subject-agnostic baseline across most held-out subjects.
While performance varies across individuals, \method matches or exceeds the baseline for some subjects (e.g., Subjects 3 and 9), indicating that the shared representations capture meaningful and transferable EEG structure even without subject-specific adaptation.

We further assess the transferability of subject-specific adapters by applying adapters trained on a source subject to unseen target subjects.
Transferred adapters consistently outperform both zero-shot \method and the subject-agnostic baseline for many subject pairs.
Notably, certain adapters yield substantial improvements for specific held-out subjects (e.g., Adapter 1 for Subject 3 and Adapter 8 for Subject 4), highlighting the presence of cross-subject structure that can be exploited without retraining the full model.

Overall, these results demonstrate that \method supports zero-shot generalization for cross-subject adaptation, while the results of transferred adapters indicate a few-shot personalization without retraining or initialization for LoRA fine-tuning.

\begin{table*}[ht]
  \caption{\textbf{Numerical Results.} Performance comparison across model variants: a subject-agnostic model (no personalization), subject-specific models (trained independently per subject), LoRA (independent low-rank model per subject), and {\method}. }
  \label{tab:results}
  \centering
  \resizebox{\textwidth}{!}{
  \begin{tabular}{lcccc}
    \toprule        
    \textbf{Method}                           & \textbf{Total}          &  \textbf{Active}  & \textbf{BCI Comp. IV2a} & \textbf{BCI Comp. IV2b}\\
                                              & \textbf{\#  Params}     &  \textbf{\# Params}        & \textbf{(4 Classes)} & \textbf{(2 Classes)}  \\
        \midrule
    EEGNeX Subject-Agnostic \cite{Chen2024EEGNeX}  &\phantom{1}55 972   & 55 972 & 55.00\% $\pm$ 0.66\% & 74.28\% $\pm$ 0.39\% \\
    EEGNeX Subject-Specific                     &\phantom{1}55 972      & 55 972 &  59.36\% $\pm$ 14.42\%  &  75.92\% $\pm$ 14.32\% \\
    EEGNex    LoRA           &\phantom{1}81 124      & 10 980 & 53.60\% $\pm$ 2.91\% & 69.64\% $\pm$ 0.27\% \\
    \textbf{EEGNeX {\method}}                   & 134 884   & 64 740    & \textbf{64.72\% $\pm$ 0.56\%} & \textbf{76.48\% $\pm$ 0.25\%} \\
    \midrule
    PBT Subject-Agnostic \cite{Klein2025PBT}    & \phantom{1} 867 460   & 867 460 &  50.82\% $\pm$ 0.85\% & 75.49\% $\pm$ 1.10\% \\
    PBT Subject-Specific                        & \phantom{1} 867 460   & 867 460&  44.16\% $\pm$ 12.19\% & 75.63\% $\pm$ 14.23\% \\
    PBT LoRA                 & \phantom{1} 680 736   & 135 712     & 25.81\% $\pm$ 0.52\% & 53.70\% $\pm$ 0.80\% \\
    \textbf{PBT {\method}}                      & 1 480 612             & 935 588     & \textbf{54.51\% $\pm$ 0.30\%} & \textbf{76.36\% $\pm$ 0.23\%} \\
    \bottomrule
  \end{tabular}
  }
\end{table*}

\subsection{Representation Structure and Subject Conditioning}
Figure~\ref{fig:tSNE} presents t-SNE visualizations \cite{vandermaatent} of the learned representations to analyze the effect of subject conditioning on the latent space. 
We compare embeddings derived from the general weights $W_{\text{general}}$, the subject-specific weights $W_s$, and the fused representations produced by the complete \method layer. 
All embeddings are extracted from the classification token predicted by the Patched Brain Transformer, using a linear classification head for visualization.
The top row of Figure~\ref{fig:tSNE} colors embeddings by subject identity, while the bottom row colors the same points by ground-truth target labels. 
Representations obtained from the general weights exhibit substantial overlap across subjects, 
with limited separation by either subject or class. 
Subject-specific representations show clearer grouping by subject identity, but this structure does not consistently align with the target labels.
In contrast, the fused representations produced by the full \method display well-defined sub-clusters within each subject’s embedding space. 
These sub-clusters correspond closely to the ground-truth target labels, as evidenced by the alignment between the top-right and bottom-right panels of Figure~\ref{fig:tSNE}. This indicates that the proposed model learns task-relevant, discriminative features while preserving subject-dependent structure in the representation space.
Overall, these visualizations provide qualitative evidence that \method improves the organization of the latent space by reducing inter-subject variability while enhancing class separability within each subject.

\subsection{Application to Motor Imagery EEG Decoding}
We evaluate the \method\ on motor imagery EEG decoding by integrating it into two distinct architectures: 
EEGNeX (see Sec.~\ref{sec:eegnex}) and the Patched Brain Transformer (see Sec.~\ref{sec:pbt}). 
The following experiment is conducted on the BCI Competition IV-2a and IV-2b datasets (see Sec.~\ref{sec:motorimaginarytask}). 
All numerical results are reported as the mean and standard deviation of three reproducible runs with random seeds $\{1,2,3\}$.
Further details on the experimental setup are provided in Appendix~\ref{app:ExperimentSetup}.

First, to establish the necessity of subject-specific adaptation, we compare against a \textbf{Subject-Agnostic Model}. 
This baseline is a standard architecture trained on data collected from all subjects, without personalization. 
Our method surpasses this baseline, confirming that a one-size-fits-all approach is insufficient for high performance in EEG decoding, Table~\ref{tab:results}.

Next, we demonstrate the benefit of knowledge sharing by comparing our method against the common practice of training fully independent \textbf{Subject-Specific Models}. This baseline trains a separate model from scratch for each individual. In Table~\ref{tab:results}, we report the mean performance across all subjects (the complete results are provided in Table~\ref{tab:resultsperSub}). {\method} consistently outperforms the average of those independently trained models. This finding highlights that the shared backbone ($W_{\text{general}}$) enables effective knowledge transfer across subjects, yielding better generalization than isolated training on limited single-subject datasets.

Finally, to validate our specific architectural design, we introduce a crucial parameter control: LoRA EEGNex. 
For each subject, a separate low-rank model is trained from scratch.
This baseline allows us to test if the performance gain is simply 
an artifact of having a certain number of trainable parameters per subject. 
{\method}'s superiority in this comparison is a critical finding. 
It proves that our hybrid architecture is more effective than training isolated, parameter-efficient models. 
This validates that the structure of our approach is important for its performance.

\section{Conclusion}
We presented the Subject-Specific Low-Rank Adapter, a method for multi-subject neural signal decoding that explicitly separates shared and subject-specific sources of variation. 
The method integrates into existing architectures with minimal modification, requiring only replacement of standard linear or convolutional layers. 

{\method} addresses a fundamental obstacle to scaling foundation models for brain decoding: inter-subject distribution shift. 
The method decomposes network weights into a general component that is learned across all subjects and lightweight, low-rank subject-specific adapters.
Our experiments across MEG speech perception and EEG motor imagery tasks demonstrate that {\method} consistently outperforms both subject-agnostic baselines and independently trained subject-specific models. 
In speech decoding, {\method} achieves superior accuracy while reducing subject-specific parameters by half compared to previous studies. 
In the motor imagery classification task, it improves performance for both CNN and transformer architectures across BCI datasets. 

The visualization of learned representations confirms that {\method} organizes the latent space by reducing inter-subject variability while enhancing class separability within each subject.

More broadly, the design underlying {\method} may extend to other multi-source learning problems characterized by systematic distribution shifts across data sources.

\section{Limitations}

By design, {\method} requires subject identity at both training and inference time. 
This prevents direct application to entirely unseen subjects without adaptation. 
Although our transferred adapter experiments suggest that cross-subject structure exists, 
leveraging this for true zero-shot generalization to new subjects remains future work.

The selection of the appropriate rank $r$ and the scaling factor $\alpha$ requires parameter tuning, 
and optimal values can vary between tasks and architectures. 
Our experiments suggest $\alpha \approx 3r$ works well, but an exact selection criterion remains an open question.

Our evaluation focuses on datasets with 9–27 subjects. 
Although the results are consistent across different modalities and architectures, 
validating {\method} on a larger scale, such as with hundreds of subjects, 
would further strengthen the motivation for the foundation model.

Finally, the total parameter count increases with the number of subjects, 
although modestly due to the low-rank constraint. 
For applications involving a substantial number of subjects, it may be essential 
to employ strategies like sharing adapters among similar subjects 
or utilizing hierarchical adapter structures

\newpage

\section*{Impact Statement}
This work aims to advance the fields of machine learning and neuroscience, in particular by 
improving the generalization of models across individuals in neural signal decoding. The proposed 
methods are intended to increase robustness and sample efficiency in learning from human brain 
data, which benefit research in brain–computer interfaces and cognitive neuroscience. The work 
does not introduce new data collections, and we do not anticipate negative ethical or societal 
consequences beyond those commonly associated with machine learning methods applied to brain 
signals.

\section*{Acknowledgments}
This project has received funding from the German Federal Joint Committee (Grant 01VSF23017),
from the European Regional Development Fund (grants timing Matters and IntelAlgen) under the European Union’s Horizon Europe Research and Innovation
Program, and from the German Research Foundation DFG (grants GRK 2297 and 537063406), which we gratefully acknowledge.

The work of Steffen Schotthöfer is sponsored by the Applied Mathematics Progrm at the Office of Advanced Scientific Computing Research, U.S. Department of Energy, and performed at the Oak Ridge National Laboratory, which is managed by UT-Battelle, LLC under Contract No. DE-AC05-00OR22725 with the U.S. Department of Energy. The United States Government retains and the publisher, by accepting the article for publication, acknowledges that the United States Government retains a non-exclusive, paid-up, irrevocable, world-wide license to publish or reproduce the published form of this manuscript, or allow others to do so, for United States Government purposes. The Department of Energy will provide public access to these results of federally sponsored research in accordance with the DOE Public Access Plan (http://energy.gov/downloads/doe-public-access-plan).
This manuscript has been authored by UT-Battelle, LLC under Contract No.~DE-AC05-00OR22725 with the U.S.~Department of Energy. The United States Government retains and the publisher, by accepting the article for publication, acknowledges that the United States Government retains a non-exclusive, paid-up, irrevocable, world-wide license to publish or reproduce the published form of this manuscript, or allow others to do so, for United States Government purposes. The Department of Energy will provide public access to these results of federally sponsored research in accordance with the DOE Public Access Plan(\url{http://energy.gov/downloads/doe-public-access-plan}).

\bibliography{references}
\bibliographystyle{icml2026}

\newpage
\appendix
\onecolumn

\section{Extension to Convolutional Layers}\label{app:subConv}
While the formulation in Equation~\ref{eq:subject-layer} is presented for linear transformations, it can be seamlessly extended to convolutional layers. A convolution operation is itself a linear transformation, where the weight matrix $W$ is replaced by a convolutional kernel $W$. The \method layer can thus be expressed as:
\begin{align}
\bar{X} = \sigma\left(X * \mathcal{W}_{\text{general}} +  \textstyle\sum_{s=1}^{N} M_s \cdot (X * \mathcal{W}_{s}) \right), \label{eq:conv-subject-layer}
\end{align}
where $*$ denotes the convolution operation.

To implement the low-rank adaptation (Section~\ref{sec:lowRank}) for the subject-specific kernels, we decompose the operation $X*W_{s}$ into two sequential convolutions, effectively factorizing the kernel itself. 
This is analogous to the matrix decomposition in Equation~\ref{eq:lowRankSub}. 
Specifically, the subject-specific update is computed by:
\begin{align}
\mathcal{W}_{s} := A_s B_s,
\end{align}
where $A_s$ and $B_s$ are the low-rank convolutional kernels. 
The first kernel, $A_s$, reduces the channel dimension from $C_{\text{in}}$ to the rank $r$ while preserving the spatial kernel size. 
The second kernel, $B_s$, is a $1\times1$ convolution that projects the features from the rank $r$ space back to the $C_{\text{out}}$ channel dimension. 
Initialization is equal to the linear case, see Section~\ref{sec:sub_lin_layer}. This convolutional decomposition allows for parameter-efficient adaptation for each subject while maintaining the inductive biases of convolutional layers.

\section{Numerical Experiments}

\subsection{Experiment Design} \label{app:ExperimentSetup}

\subsubsection*{EEGNeX}
We used the EEGNeX implementation from the Braindecode package \cite{GramfortBraindecode, SchirrmeisterBraindecode}. In this model, we replaced all standard convolutional layers with the convolutional version of \method, (see Appendix~\ref{app:subConv}). We made an exception for the depthwise convolutional layer (DepthwiseConv2D) in Block 3 due to implementation details.

As the original hyperparameters for EEGNeX were not specified, we tuned the baseline learning rate by sweeping over the values $\{10^{-3}, \; 10^{-4}\}$. We want to remark that the difference between the numbers reported in Table~\ref{tab:results} and those in the original EEGNeX paper stems from different evaluation methods. Here, we report the numbers for cross-session analysis from a 512-timepoint long interval.

The final hyperparameters are reported in Table~\ref{tab:hyperparametersEEGNeX}:

\begin{table}[!ht]
\caption{Hyperparameters  EEGNeX.}
\label{tab:hyperparametersEEGNeX}
\centering
\begin{tabular}{ll}
\toprule
\textbf{Hyperparameter} & \textbf{Value} \\
\midrule
Band-pass filter & [8, 45]\\
Optimizer & AdamW \\
Learning rate & $1\times 10^{-3}$ \\
Weight decay & 0.01 \\
AdamW betas & (0.9, 0.999) \\
Batch size & 64 \\
Training epochs & 100\\
Dropout rate & 0.0 \\
Activation function & ELU \\
Rank adapters & 4\\
LoRA $\alpha$ & 1\\
\bottomrule
\end{tabular}
\end{table}

\subsubsection*{Patched Brain Transformer}
For the PBT, we used the official GitHub repository accompanying the publication and replaced all linear layers with \method. The only exception was for the visualization in Figure~\ref{fig:tSNE}, where we kept the final linear layer to improve clarity.

We used the same hyperparameters as reported in the original PBT paper for all runs (Table~\ref{tab:hyperparametersPBT}). For our method's specific hyperparameters, we set the rank $r=8$ and $\alpha=1$.

\begin{table}[h]
\caption{Hyperparameters PBT.}
\label{tab:hyperparametersPBT}
\centering
\begin{tabular}{ll}
\toprule
\textbf{Hyperparameter} & \textbf{Value} \\
\midrule
Band-pass filter & [8, 45]\\
Optimizer & AdamW \\
Learning rate & $3 \times 10^{-4}$ \\
Weight decay & 0.01 \\
AdamW betas &(0.9, 0.95) \\
Batch size & 128 \\
Training epochs & 1200\\
Warmup epochs & 150 \\
Dropout rate & 0.1 \\
Activation function & GELU \\
Rank adapters & 8\\
LoRA $\alpha$ & 1\\
Data augmentation & Time shifts\\
\bottomrule
\end{tabular}
\end{table}

\subsubsection*{BrainMagick}
For BrainMagick, we used the official GitHub repository accompanying the publication and replaced the subject-specific layer with {\method}. We used the same hyperparameters as reported in the original paper for all runs. Specific hyperparameters of our method, respective rank $r$ and $\alpha$ are reported in Table~\ref{tab:BrainMagick}.

\begin{table}[h!]
\caption{Hyperparameters BrainMagick.}
\label{tab:hyperparametersBrainMagick}
\centering
\begin{tabular}{ll}
\toprule
\textbf{Hyperparameter} & \textbf{Value} \\
\midrule
Signal processing & High-pass filter, downsampling\\
Baseline correction & 0.5 s average subtraction\\
Normalization & Robust scaler\\
Speech representation & Wav2vec 2.0 (pretrained)\\
Subject embedding & Yes\\
Convolutional blocks & Stacked CNN architecture\\
Activation function & GELU\\
Loss function & Contrastive (CLIP-style)\\
Optimizer & Adam\\
Batch size & 256\\
Learning rate & $3\times 10^{-4}$\\
Training epochs & Early stopping ($\sim$50\% of test set)\\
Dropout & Spatial attention dropout\\
Segment duration & 3 seconds\\
\bottomrule
\end{tabular}
\end{table}

\newpage

\section{Detailed Numerical Results for Subject-Specific Modeling}

\begin{table}[!ht]
  \caption{Subject-Specific Numerical results for EEGNeX and PBT on individual subjects.}
  \label{tab:resultsperSub}
  \centering
  \begin{tabular}{lllll}
  \toprule
    \textbf{Subject} & \multicolumn{2}{c}{\textbf{EEGNeX}} & \multicolumn{2}{c}{\textbf{PBT}}\\
    \cmidrule(lr){2-3} \cmidrule(lr){4-5} 
         & BCI Comp. IV2a    &   BCI Comp. IV2b & BCI Comp. IV2a    &   BCI Comp. IV2b\\
    \midrule
    1 & 67.82\% $\pm$ 6.25\% & 72.50\% $\pm$ 1.84\% &  56.60 \% $\pm$ 1.85\%      & 71.35\% $\pm$ 1.28\% \\
    2 & 46.53\% $\pm$ 1.70\% & 52.86\% $\pm$ 0.77\% &  25.23 \% $\pm$ 1.31 \%     & 53.10\% $\pm$ 1.02\% \\
    3 & 77.08\% $\pm$ 3.27\% & 51.56\% $\pm$ 1.99\% &  55.90 \% $\pm$ 3.07 \%    & 50.73\% $\pm$ 0.15\% \\
    4 & 53.01\% $\pm$ 1.80\% & 94.27\% $\pm$ 0.53\% &  39.00 \% $\pm$ 1.15 \%    & 95.31\% $\pm$ 1.11\% \\
    5 & 36.69\% $\pm$ 3.38\% & 86.25\% $\pm$ 1.35\% &  26.97 \% $\pm$ 0.59 \%    & 86.98\% $\pm$ 1.62\% \\
    6 & 40.97\% $\pm$ 1.58\% & 81.77\% $\pm$ 2.32\% &  34.72 \% $\pm$ 2.79 \%    & 77.81\% $\pm$ 1.02\% \\
    7 & 67.94\% $\pm$ 2.31\% & 70.94\% $\pm$ 2.04\% &  47.80 \% $\pm$ 7.45 \%    & 78.52\% $\pm$ 0.72\% \\
    8 & 68.29\% $\pm$ 0.71\% & 86.77\% $\pm$ 1.49\% &  55.79 \% $\pm$ 3.38 \%    & 86.46\% $\pm$ 0.64\% \\
    9 & 75.93\% $\pm$ 0.43\% & 83.85\% $\pm$ 1.98\% &  55.44 \% $\pm$ 0.59 \%    & 83.02\% $\pm$ 1.64\% \\

    \bottomrule
  \end{tabular}
\end{table}

\section{Software and Data}\label{sec:softwareData}
Upon acceptance, a PyTorch implementation of \method will be released on GitHub, together with a model wrapper that allows the layer to be easily incorporated into existing architectures, including models from the Braindecode package \cite{braindecode_code}. 
For the baseline model implementations used in our experiments, we utilized: the EEGNeX implementation from the Braindecode package \cite{braindecode_code}, the PBT implementation from the paper's GitHub repository \cite{pbt_code}, and the BrainMagick implementation from the paper's GitHub repository  \cite{brainmagick_code}.

All datasets used in this study are publicly accessible, either directly from OSF (\url{https://osf.io/ag3kj/}) or via the BCI Horizon 2020 repository (\url{http://bnci-horizon-2020.eu/database/data-sets}). Experiments on BCI datasets are conducted within the MOABB-Package \cite{chevallierMOABB, AristimunhaMOABB}.

\section{Compute Resources}
The experiments for EEGNeX and PBT were performed on a single NVIDIA Tesla V100 GPU using PyTorch. The Brain Magick experiments were performed on single NVIDIA H100 GPU.


\end{document}